\documentclass[conference]{IEEEtran}
\IEEEoverridecommandlockouts
\usepackage{cite}
\usepackage{amsmath,amssymb,amsfonts}
\usepackage{algorithmic}
\usepackage{graphicx}
\usepackage{textcomp}
\usepackage{xcolor}
\def\BibTeX{{\rm B\kern-.05em{\sc i\kern-.025em b}\kern-.08em
    T\kern-.1667em\lower.7ex\hbox{E}\kern-.125emX}}
\usepackage{bm}
\usepackage{fancyhdr}
\fancypagestyle{withfooter}{
  
  \fancyfoot[C]{\footnotesize Accepted to the IEEE ICRA Workshop on Field Robotics 2024 \\ Distributed under the terms of the Creative Commons Attribution License (CC BY 4.0)}
}

\usepackage{subcaption}

\begin{document}

\title{Aquaculture field robotics: \\Applications, lessons learned and future prospects}

\makeatletter
\newcommand{\linebreakand}{%
  \end{@IEEEauthorhalign}
  \hfill\mbox{}\par
  \mbox{}\hfill\begin{@IEEEauthorhalign}
}
\makeatother

\author{\IEEEauthorblockN{Herman B. Amundsen}
\IEEEauthorblockA{\textit{Dept. Aquaculture Technology} \\
\textit{SINTEF Ocean}\\
Trondheim, Norway \\
herman.biorn.amundsen@sintef.no}
\and
\IEEEauthorblockN{Marios Xanthidis}
\IEEEauthorblockA{\textit{Dept. Aquaculture Technology} \\
\textit{SINTEF Ocean}\\
Trondheim, Norway \\
marios.xanthidis@sintef.no}
\and
\IEEEauthorblockN{Martin Føre}
\IEEEauthorblockA{\textit{Dept. Engineering Cybernetics} \\
\textit{NTNU}\\
Trondheim, Norway \\
martin.fore@ntnu.no}
\linebreakand
\IEEEauthorblockN{Sveinung J. Ohrem}
\IEEEauthorblockA{\textit{Dept. Aquaculture Technology} \\
\textit{SINTEF Ocean}\\
Trondheim, Norway \\
sveinung.ohrem@sintef.no}
\and
\IEEEauthorblockN{Eleni Kelasidi}
\IEEEauthorblockA{\textit{Dept. Aquaculture Technology} \\
\textit{SINTEF Ocean}\\
Trondheim, Norway \\
eleni.kelasidi@sintef.no}
}
\maketitle

\thispagestyle{withfooter}
\pagestyle{withfooter}

\begin{abstract}
Aquaculture is a big marine industry and contributes to securing global food demands.
Underwater vehicles such as remotely operated vehicles (ROVs) are commonly used for inspection, maintenance, and intervention (IMR) tasks in fish farms.
However, underwater vehicle operations in aquaculture face several unique and demanding challenges, such as navigation in dynamically changing environments with time-varying sealoads and poor hydroacoustic sensor capabilities, challenges yet to be properly addressed in research.
This paper will present various endeavors to address these questions and improve the overall autonomy level in aquaculture robotics, with a focus on field experiments.
We will also discuss lessons learned during field trials and potential future prospects in aquaculture robotics.
\end{abstract}

\begin{IEEEkeywords}
Aquaculture, Marine Robotics, Field Robotics, Autonomous systems
\end{IEEEkeywords}

\section{Introduction}


Aquaculture is an important source of protein, and will likely play an even more important role in securing global food demands going forward. 
In 2020, global aquaculture reached a production of 122.6 million tons, with a total value of 281.5 billion USD~\cite{fao:2022}.
In Norway, aquaculture has been an industrial success story; from its humble beginnings in the 1970s, it has grown to be Norway's second-largest export industry, exceeded only by oil and gas.
Atlantic salmon (\textit{Salmo salar}) is the dominating species, and most of the production is conducted in floating fish farms along the Norwegian coast.

Farms consist of net cages, which are flexible structures that can deform with waves and currents. 
A net cage typically consists of a net enclosure that is suspended from a floating collar and whose lower edge is attached to bottom weights or a bottom ring to maintain a sufficient volume~\cite{faltinsen:2018:fish_farms}.
Further components in fish farms include mooring systems, automatic feeding systems, and various instrumentation.
While dimensions vary, a typical fish farm can consist of more than 10 net cages, all measuring 50~m in diameter and 30~m in depth, and containing up to 200,000 individuals~\cite{Fore:2024:digital_twins}. 

Fish farms are typically situated in sheltered coastal waters.
However, due to a lack of available sites and because of environmental concerns, there is a trend of moving fish farms further offshore where the facilities are more exposed to weather and sealoads~\cite{Bjelland:2015:exposed,morro:2021:exposed}. 

\begin{figure}
    \centering
    \includegraphics[width=0.9\columnwidth]{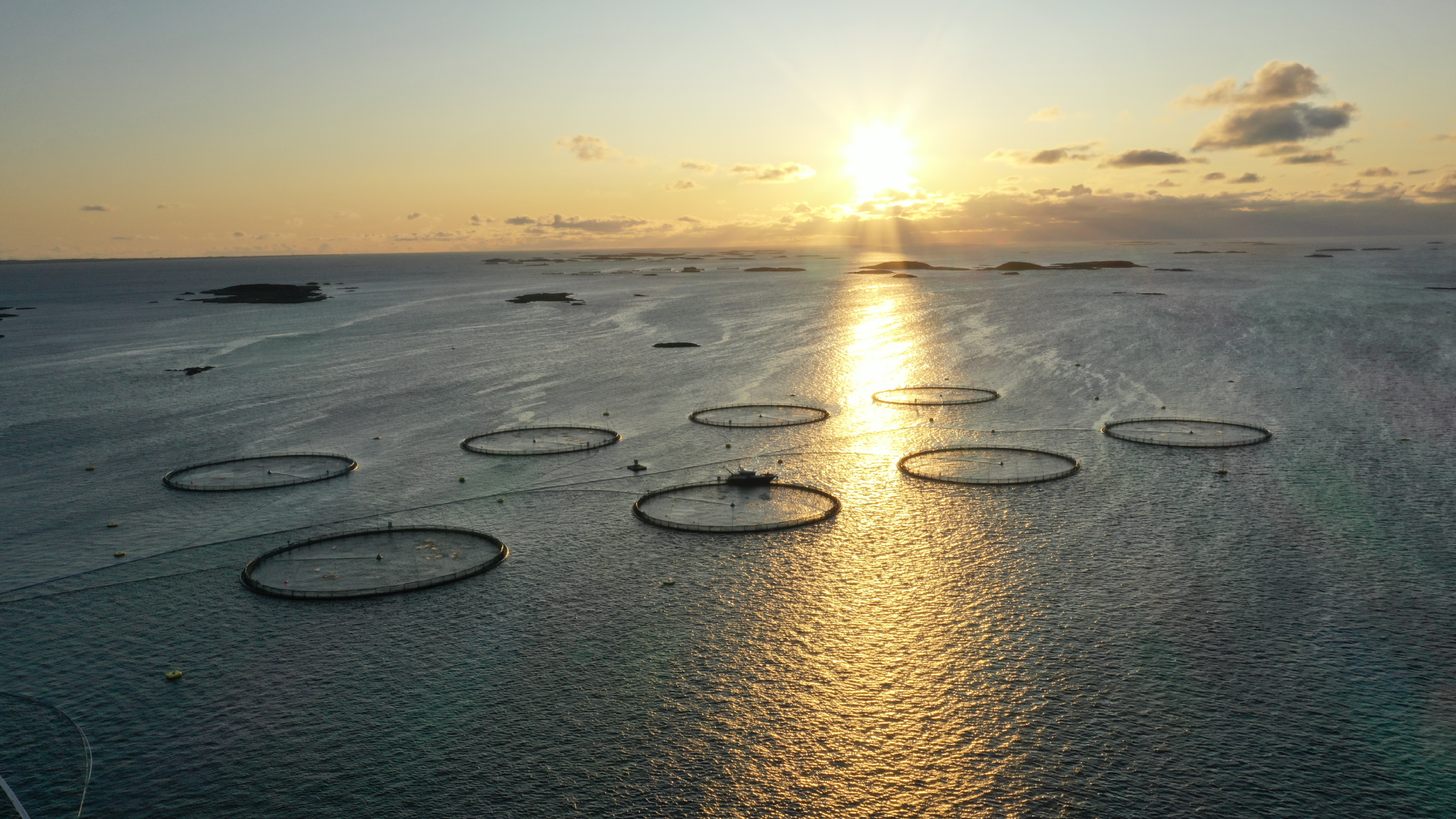}
    \caption{SINTEF ACE is a full-scale aquaculture laboratory consisting of several fish farms at various locations along the Norwegian coast. Pictured is the facility at Rataren, Frøya.}
    \label{fig:ACE}
\end{figure}

Traditionally, fish farms have been dependent on divers for monitoring and intervention operations inside the cages.
As diving is associated with risk and because of stricter governmental regulations, unmanned underwater vehicles (UUVs) such as ROVs have been replacing divers for the last decades and are now indispensable tools for Norwegian fish farmers.  
Common ROV operations include cleaning the nets from biofouling~\cite{Bannister2019}, mooring line inspections, and fish monitoring~\cite{kelasidisvendsen2022}.
The most important operation, however, is likely to inspect the nets for holes and structural failures, as such deficiencies can lead to the escape of farmed fish, which both represent a production loss and an environmental concern, as escapees may impact wild fish populations~\cite{MoeFore2021}.
Net inspections are therefore performed on a regular basis.
Since the complex mooring systems outside the net cages increase the risk of tether entanglement, operations are usually performed inside the net cages~\cite{Rundtop2016}.
The pilot typically steers the ROV based on the video from a forward-looking camera and telemetry such as depth and heading readings.
Apart from depth and heading hold, operations are without automatic control.

Motivated by the need to reduce costs, mitigate risk from human errors, and increase the weather window when moving further offshore, there is currently a scientific effort to increase the autonomy level in ROV operations~\cite{kelasidisvendsen2022}. 
This aligns well with the concept of precision fish farming (PFF)~\cite{PrecisionFishFarming}, an ongoing effort to move the operational principles of aquaculture from manual operations and experiences-based reasoning, to autonomous operations, objective data interpretation, and decision support systems (Fig.~\ref{fig:pff}).
Research efforts include autonomous net inspection by using forward-looking sensors~\cite{Amundsen2022,Bjerkeng2021,Karlsen2021,Cardaillac2023} and camera~\cite{Livanos2018,Lin:2020:net_inspection_osv,Duecker:2020:net_inspection,Zhao:2020:net_damage,betancourt:2020:net_inspection,Schellewald:2021:net_cage_pose,Madshaven:2022:hole_detection,akram:2022}, autonomous navigating by installing instrumentation on the net cages~\cite{Arnesen2018,kelasidi2022}, and the use of autonomous net-cleaning robots that crawl along the net surface~\cite{Ohrem2021,Skaldebo:2023:remora}.

\begin{figure}
    \centering
    \includegraphics[width=0.8\columnwidth]{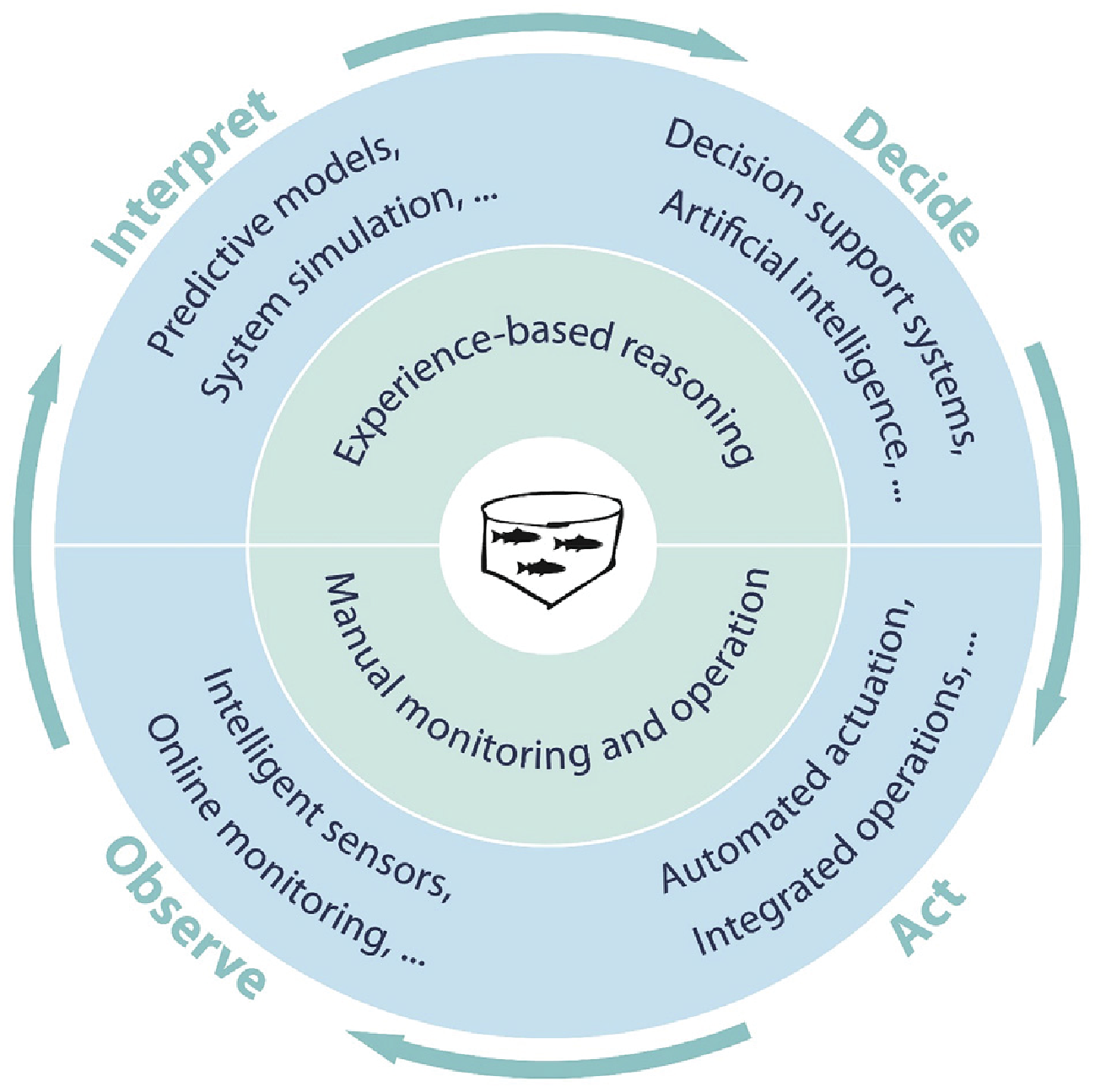}
    \caption{Precision fish farming (PFF) aims at moving the operational principles of aquaculture from the current standard of manual operations and experience-based reasoning to a more control-oriented approach~\cite{PrecisionFishFarming}.}
    \label{fig:pff}
\end{figure}

In the underwater robotic domain, aquaculture represents an especially demanding environment with many unique challenges that are non-trivial to solve. 
Operations take place in the wave-zone~\cite{Lader2017} and in irregular ocean currents induced by the dense biomass~\cite{Gansel2014}, which greatly affects control performance. 
Further, sensor capabilities are often degraded, as hydroacoustic sensors suffer from scattering and multipath propagation effects due to the air-filled swim-bladder of the fish~\cite{Rundtop2016}, and cameras are affected by frequent occlusions~\cite{Schellewald:2021:net_cage_pose}.
Finally, operations take place in a highly dynamic and cluttered environment consisting of flexible structures and dense biomass, where safety is critical as collisions can harm the vehicle, the fish, and the net structure. 

In this article, we will present results from a decade of field experiments in aquaculture robotics. 
These field experiments have been conducted in SINTEF ACE~\cite{sintef-ace-website}, an industrial-scale aquaculture laboratory consisting of several fish farms dedicated to aquaculture research and development (Fig.~\ref{fig:ACE}). 
We will discuss lessons we have learned over the years and future prospects, with the aim that this paper can contribute to further interest in aquaculture robotics, a domain in which autonomy holds huge potential. 
The paper is organized as follows:
In Section~\ref{sec:topics}, we present results and experiences from field experiments divided into four topics; localization, planning, control, and fish monitoring. 
Then, lessons learned are presented in Section~\ref{sec:lessons}, while Section~\ref{sec:future} discusses future prospects.
Finally, the paper is concluded in Section~\ref{sec:conclusion}. 

\section{Field work}\label{sec:topics}
\subsection{Localization, perception, and mapping}\label{topic:localization}
A fundamental challenge of most autonomous robotic systems is that of localization; without knowing the states of the robot, one cannot make informed decisions on future actions. 
In aquaculture, this is further complicated by the fact that the surroundings of the robot are in a constant state of motion and that most operational objectives are usually defined relative to the moving net structure~\cite{Amundsen2022}.
It is therefore not enough to determine a georeferenced position of the robot, one must also determine the relative position between the robot and the net structure. 
This challenge has generated different methods that can largely be split into a local and global category. 
Local approaches rely on a forward-looking sensor or camera to estimate the robot's pose relative to a local section of the net right in front of the robot, and in turn, make decisions based on this estimate.
In contrast, global approaches are based upon concurrently estimating both the robot position and the net shape, and then navigating using these estimations. 

\subsubsection{Evaluation of hydroacoustic instrumentation}

\begin{figure}
    \centering
    \includegraphics[trim={16cm 8cm 0 8cm},clip,width=0.8\columnwidth]{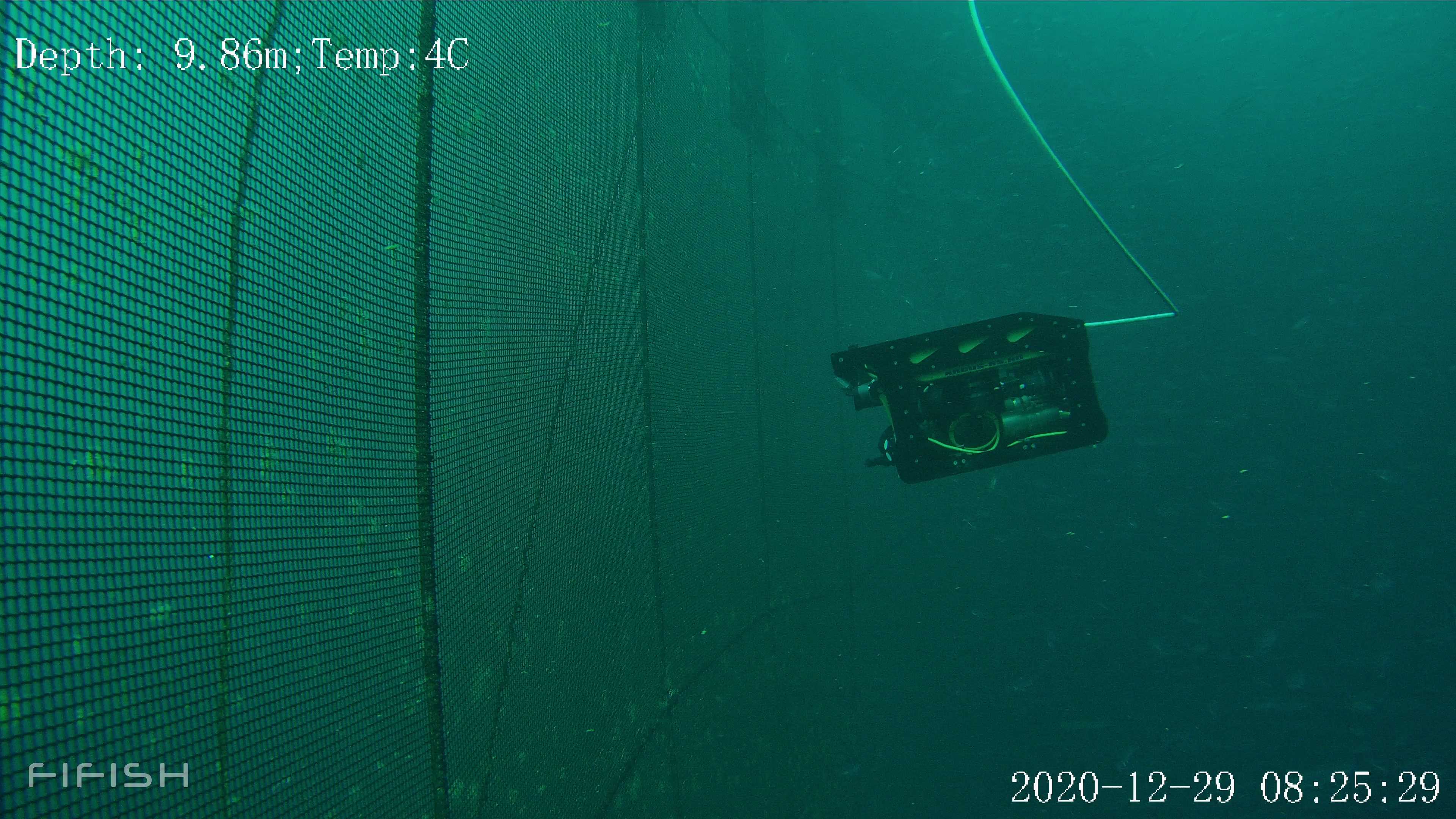}
    \caption{ROVs are commonly used for net inspections in aquaculture. 
    The image is from an experiment in SINTEF ACE where an ROV is performing autonomous net inspection using a DVL~\cite{Amundsen2022}.}
    \label{fig:dvl}
\end{figure}

Our lab's first endeavor in localization for aquaculture robotics was an experimental evaluation of hydroacoustic instrumentation in fish farms~\cite{Rundtop2016}.  
The first part of the experiment was an evaluation of an ultrashort baseline (USBL) positioning system.
A Sonardyne Scout Plus system was used in the experiments, and the performance was evaluated by analyzing the position measurements of a transponder that was placed on various sections of a net cage or attached to an ROV performing various maneuvers.
The USBL measurements had an acceptable precision and update rate, and, though both the standard deviation and dropout periods of the measurements increased with denser biomass between the transceiver and the transponder, the trials suggested that acoustic positioning systems such as USBL can be part of an ROV navigation system in aquaculture. 

The second part of the trials in~\cite{Rundtop2016} featured an evaluation of a Doppler velocity logger (DVL).
In underwater navigation, DVLs have proven very useful tools for navigation; by sending hydroacoustic beams toward the seabed, it is possible to measure the linear velocity from the Doppler shift of the reflected beams.
When the seabed is in the operational range of the DVL such that the hydroacoustic beams are reflected, the sensor is considered to have a bottom-lock. 
In an aquaculture cage bottom-lock will never be achieved as the fish and cage will interfere with the hydroacoustic beams. 
Instead, the DVL was tested by mounting it forward-looking, such that the beams would be reflected by the net cage in front of the ROV, yielding a \textit{net-lock}.
In the trials, the DVL was able to provide accurate measurements when the field-of-view (FOV) was unobstructed by fish, which was largely the case when the ROV was closer than 3 meters from the net. 
Further, the trials also showed that one could use the measured length of the DVL beams to estimate the pose of the ROV relative to the net in front, which was the inspiration for our first local method.  

\begin{figure}[t]
  \centering
  \begin{subfigure}[t]{0.45\linewidth}
    \centering\includegraphics[trim={80 60 160 20},clip,width=\linewidth]{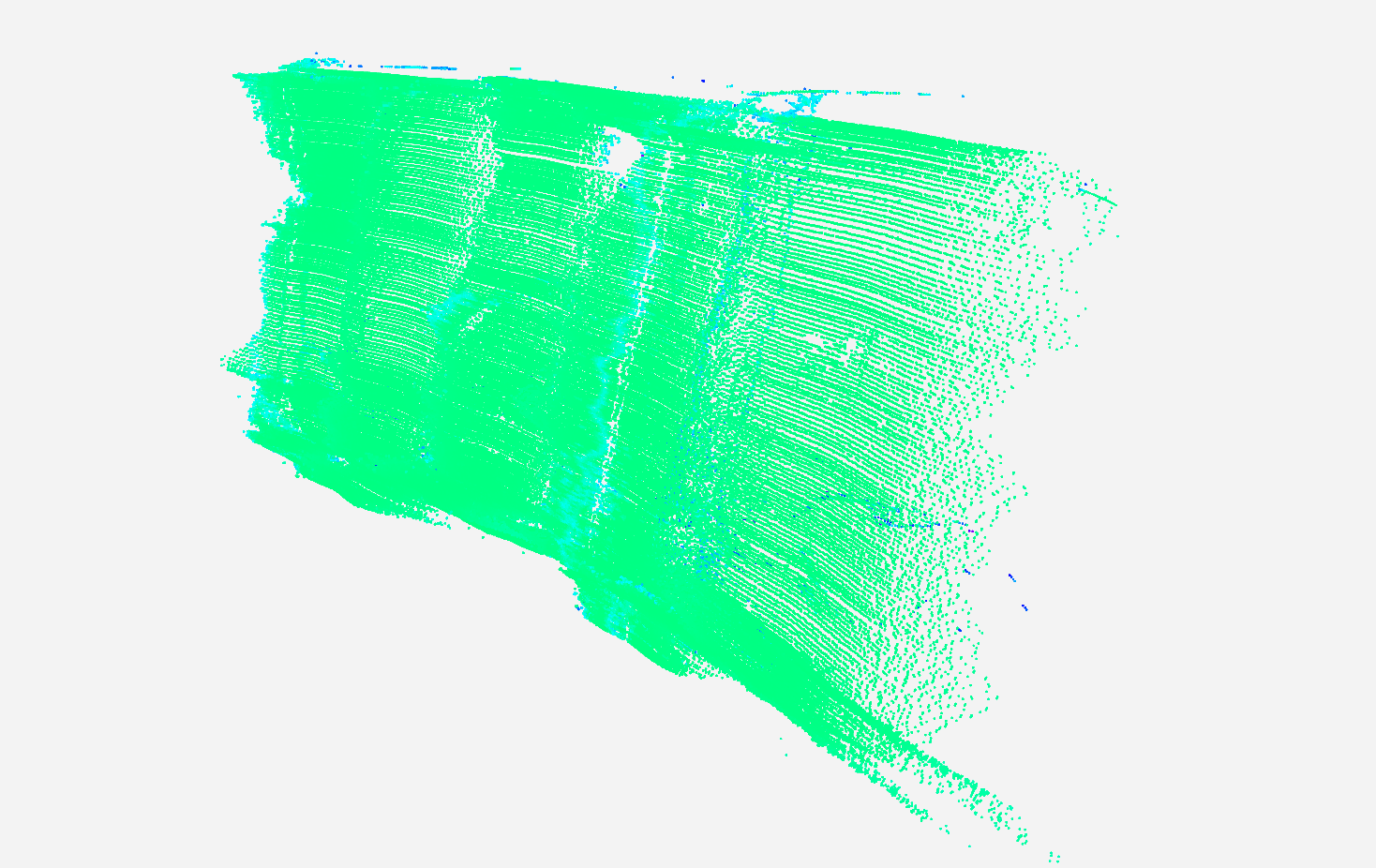}
    \caption{An acoustic point cloud of a net section.}
  \end{subfigure}%
  \hfill
  \begin{subfigure}[t]{0.45\linewidth}
    \centering\includegraphics[trim={80 60 160 20},clip,width=\linewidth]{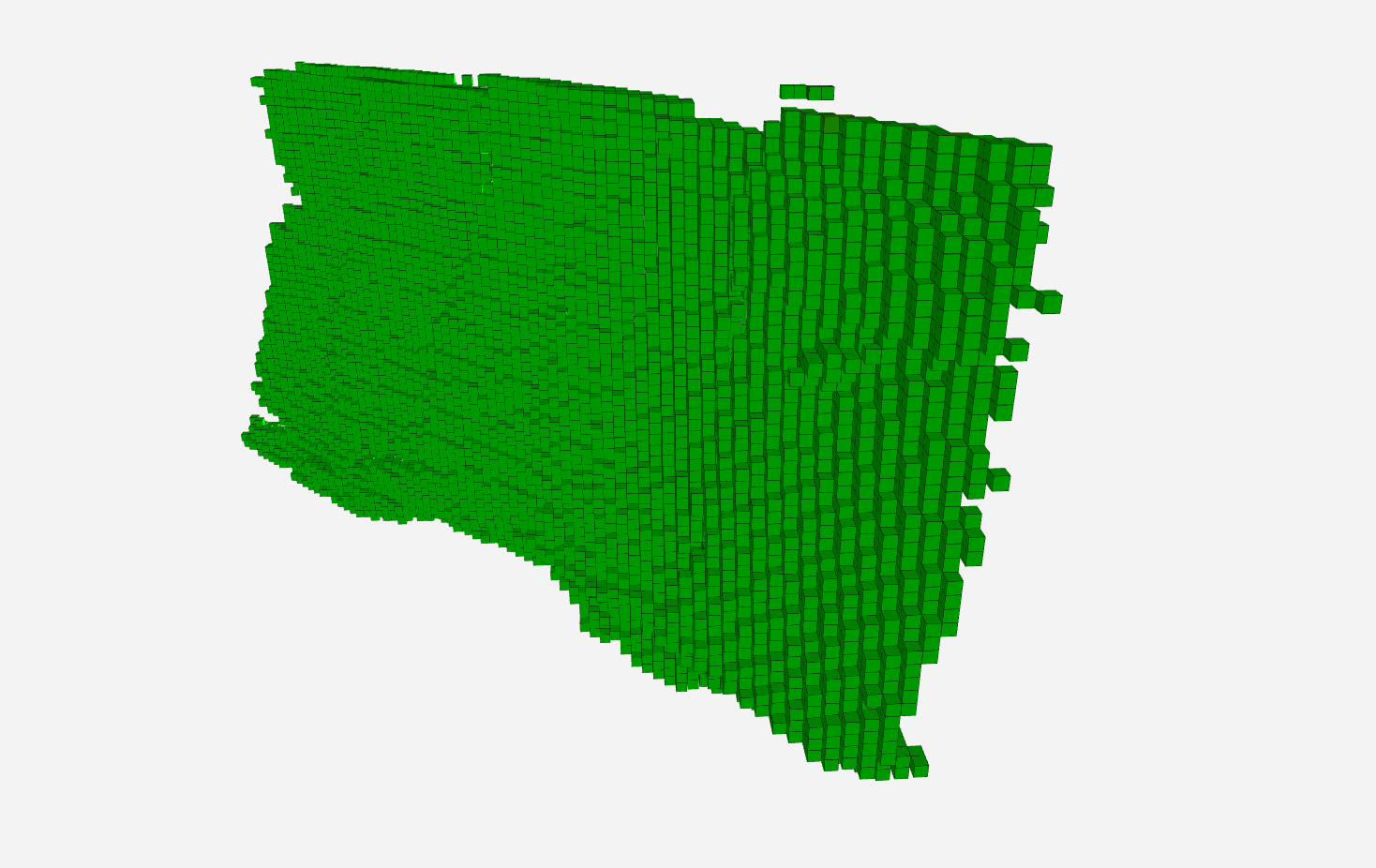}
    \caption{The point cloud converted into a voxel map.}
  \end{subfigure}%
  \newline
  \begin{subfigure}[t]{\linewidth}
    \centering\includegraphics[trim={80 60 160 20},clip,width=0.9\linewidth]{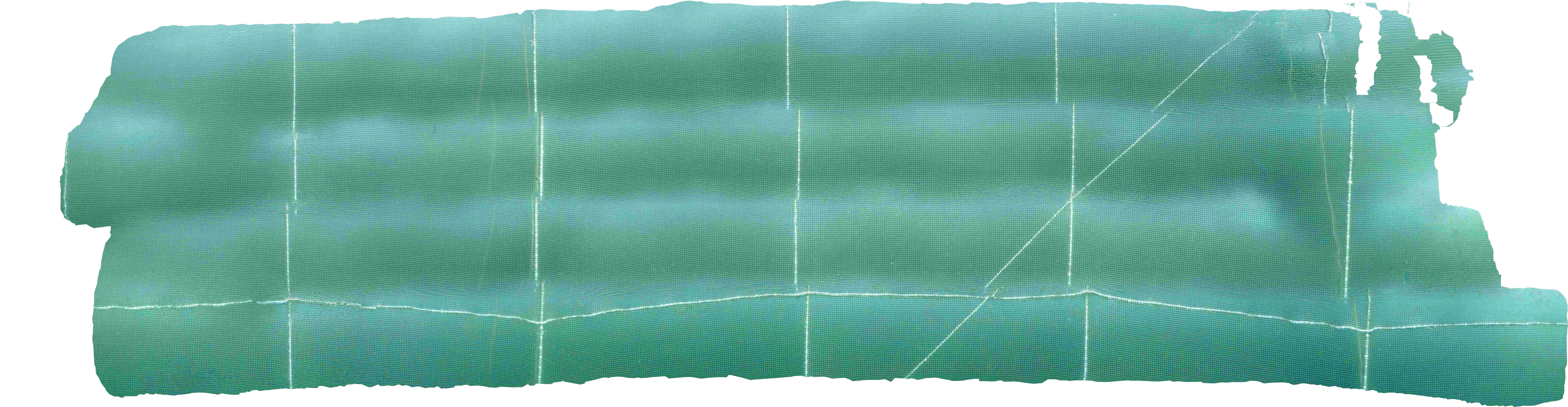}
    \caption{Orthomosaic of camera images.}
    \label{fig:orthomosaic}
  \end{subfigure}%
  \caption{Mapping during autonomous net inspection using multibeam sonar~\cite{Cardaillac2023}.}
    \label{fig:cardaillac}
\end{figure}

\subsubsection{Local positioning}
Inspired by the result of~\cite{Rundtop2016} and DVL-aided altitude control strategies~\cite{Dukan2014}, an approach for locally estimating the net-relative pose of the ROV was developed in~\cite{Amundsen2022} (Fig.~\ref{fig:dvl}).
From the measured distance of the reflected beams during net-lock, the local net section of the net in front of the ROV could be approximated as a plane. 
The plane approximation would update online as the ROV is moving, such that the approximation remained accurate regardless of displacements.
As the plane is defined in the body-fixed reference frame of the ROV, it is possible to navigate relative to the plane, even without a georeferenced state estimate. 
The method provided good accuracy and robustness when the FOV was unobstructed, but was susceptible to fish swimming between the net and the sensor, which caused outliers or loss of net-lock.


In~\cite{Cardaillac2023}, we experimented using a forward-looking multibeam sonar (FL-MBS) to approximate the net section in front of the ROV. 
The FL-MBS measurements yielded a higher resolution of the local net approximation than the DVL measurements, which could be utilized for mapping purposes as seen in Figure~\ref{fig:cardaillac}.
Collecting the measurements in a dense point map, we were able to generate a voxel map representation of the net structure.
Finally, during post-processing of the video, we were able to create an orthomosaic of the camera images.
The FL-MBS measurements also provided more robustness to occlusions than the DVL, as the measurements have more acoustic points such that outliers can easily be rejected.


\subsubsection{Global positioning}
\begin{figure}[t]
    \centering
    \includegraphics[trim={0 0 0  1cm}, clip,width=0.8\columnwidth]{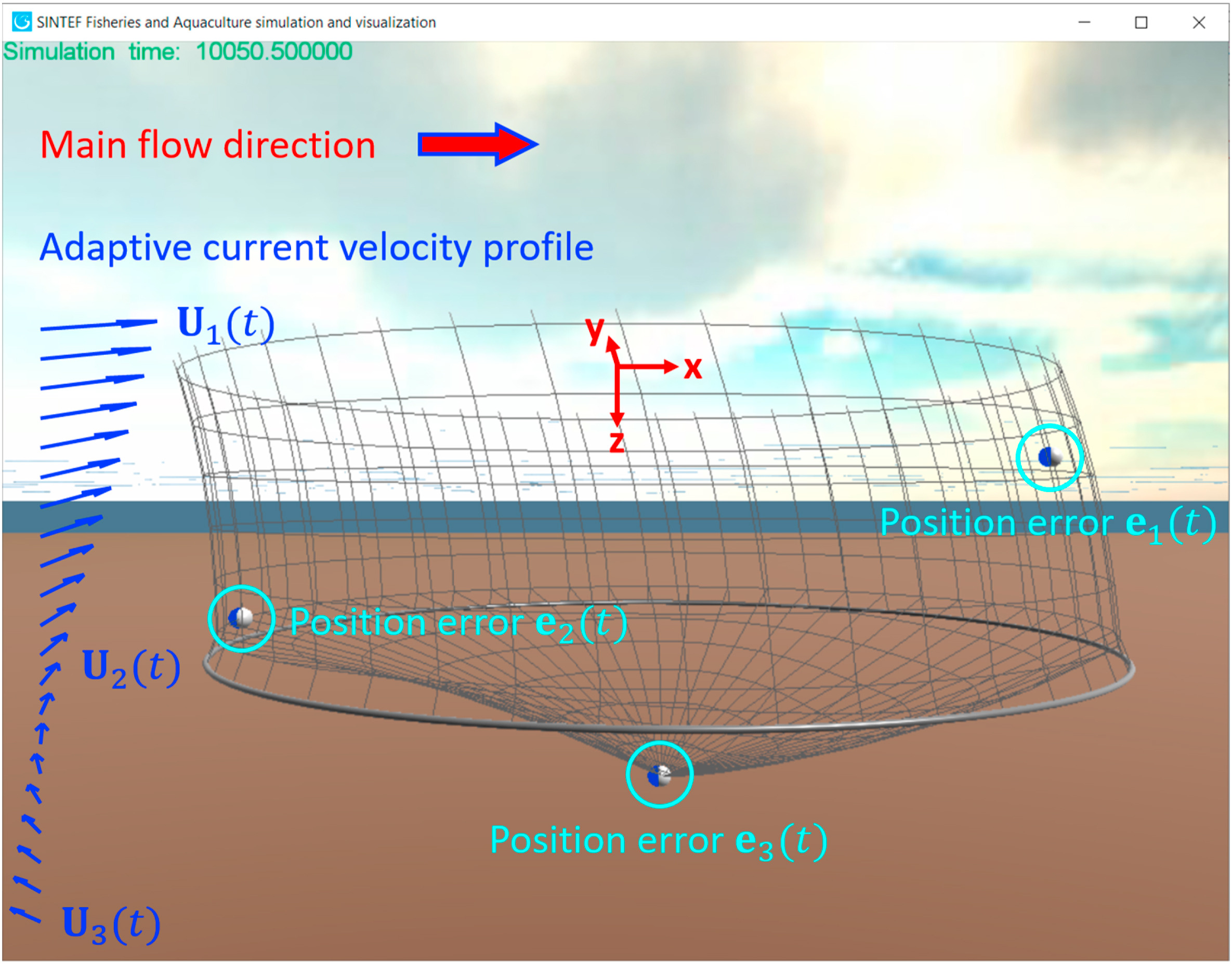}
    \caption{Online estimation of net cage deformation by using measurements from an ADCP and SBL transponders attached to the net~\cite{Su2021}.}
    \label{fig:su2021}
\end{figure}

The methods presented thus far can give a local estimate of the net-relative pose, but the estimates do not hold any validity outside of a local region. 
An alternative can be to estimate the state of the ROV and the shape of the net cage in an inertial frame, as this holds more information, making it easier to make optimal and safe plans. 
While state estimation of ROVs is a widely researched topic and can largely be solved with inertial sensors and auxiliary sensors such as DVL and USBL to avoid drift, real-time estimation of net cage structures has been less studied. 
Over the years, different models of net structure deformation have been developed~\cite{KLEBERT2015,faltinsen:2018:fish_farms,Su2021}, but have either been used in simulations with known current velocities or validated by post-processing experimental data.  
Without any measurement, the estimation error of purely mathematical models is unbounded.

In~\cite{kelasidi2022}, a method for online estimation of net structure deformation was developed. 
To bound the estimation error, three short baseline (SBL) transponders were installed at different positions on the net structure as seen in Fig.~\ref{fig:su2021}. 
From their position measurements and measurements from an adaptive current velocity profiler (ADCP), the model of~\cite{Su2021} was used to extrapolate the full net structure.
Furthermore, a fourth transponder was installed on the vehicle, such that its position was measured. 
This approach provided holistic localization with more information compared to the local approaches, at the extra cost of the integration of transponders to the net structure.

\subsection{Planning and autonomy}\label{topic:planning}
Provided information on the vehicle state and its surroundings, one can plan autonomous operations. 
A typical objective for mobile robots is safely reaching a set of waypoints, which can also be applied in aquaculture robotics, for instance, if a certain region of the net cage is of special interest or inspection of instrumentation or mooring lines.
A more specialized operation is net inspection, which can either be solved by defining a dense set of waypoints or by planning trajectories relative to a local net-relative pose. 

\subsubsection{Net inspection}
The industry standard is to perform net inspection by manually piloting the ROV along vertical or horizontal segments of the net structure while keeping the net within the FOV of the camera and at a safe distance from the ROV.
Since the pilot must perform several tasks simultaneously, such as controlling the vehicle, tether management, inspecting the video feed, and keeping track of which parts of the structure have been covered, missions are prone to human errors~\cite{fiskdir:2020:rov_inspeksjon}. 

In~\cite{Amundsen2022}, the net-relative pose estimation from the DVL was used to perform semi-autonomous net inspections. 
The operator specified a desired distance, direction, and inspection speed, and, using the plane approximation of the net, desired vertical or horizontal straight-line trajectories were computed. 
Since the net approximation will update as the vehicle moves, the desired trajectory will update accordingly, essentially simplifying the mission to a set of straight-line path following segments. 
This approach was able to relieve the operator from steering the vehicle, though the operator still had to monitor the operation and change direction (up/down/starboard/port) to gain complete coverage. 

An attempt at increasing the autonomy level further was presented in~\cite{Karlsen2021}. 
The idea was based on monitoring the progress of an inspection from the depth of the ROV and the azimuth angle of the net, which can be identified from the local approximation and will be unique under the assumption that horizontal slices of the net structure maintain concavity.
To this end, a lawn-mover pattern for inspection of the net was defined at start-up, and the direction of the inspection was changed by monitoring the depth and azimuth angle. 
While the approach worked well in simulations, outliers in the azimuth angle estimations caused problems in the field trials, such that parts of the net structure were not covered during the inspections. In~\cite{Cardaillac2023}, a different attempt was made. 
Similarly to~\cite{Karlsen2021}, an inspection pattern was defined at start-up.
In contrast, the progress was monitored by using the estimated inertial position of the ROV. 
This approach proved more successful, and full coverage of the inspected sections of the net structure was achieved.

\subsubsection{Obstacle avoidance}
While the methods presented above provide methods to traverse the net at a safe distance, they do not take into regard other parts of the net cage environment, such as instrumentation, ropes and cables, feeding systems, and the biomass itself. 
Because of the safety-critical nature of aquaculture operations, it is therefore important to also incorporate more general obstacle avoidance into fully autonomous systems.
Path planning and obstacle avoidance is a widely researched topic, but there are few examples where methods have been applied to aquaculture. 

In~\cite{amundsen:2024:field_robotics}, the elastic band path planning method~\cite{Quinlan1993} was tested in an aquaculture setting.
The experiments represented one of the first times general collision-free path planning was tested in fish farming. 
In the experiments, a separate intercepting vehicle, whose position was measured with a USBL transponder, acted as a dynamic obstacle. 
The experiments showed that the method was quickly able to plan safe paths such that the ROV was able to reach waypoints in the presence of obstacles. 
In real-world scenarios, obstacles have to be detected with the sensors of the ROV, which remains future work.

A further effort to introduce safe planning with obstacle avoidance in aquaculture was the experimental testing of ResiPlan, a motion planning framework that improves safety by adaptively changing the required clearance to obstacles with errors in the path tracking performance of the vehicle, effectively taking into account control errors, uncertainty, and environmental disturbances~\cite{Xanthidis2023}. 
In particular, the required clearance increased with state uncertainty stemming from noisy and infrequent USBL measurements that would cause jumps in state estimations, thus leading to more conservative but safer paths.

\subsection{Control systems}\label{topic:control}
While control of underwater vehicles is generally challenging due to nonlinear hydrodynamics, this is even more challenging in aquaculture as operations take place in the wave zone and external disturbances are time-varying. 
We quickly learned this in practice during net inspection trials.  PID controllers proved insufficient when controlling the velocity of the vehicle, which reduced the ROV's ability to follow the desired trajectories and thus compromised safety. 
This sparked the development and testing of a set of control laws aimed at improving this performance. 
In~\cite{Ohrem2022}, an adaptive backstepping dynamic positioning (DP) controller for ROVs was developed which can estimate both the vehicle model parameters and the direction and strength of external disturbances.
This was further developed into an adaptive velocity controller, also able to measure the same parameters~\cite{Ohrem:2023:adaptive_speed}.
Finally, a generalized super-twisting sliding mode controller was developed and tested in~\cite{Haugalokken2023}, which provided robustness to external disturbances with unknown bounds. 
These efforts were important to improve our ability to control the vehicle even in the presence of harsh environmental disturbances. 

\subsection{Fish monitoring and fish-machine interaction}\label{topic:fish_interaction}
The most important component to monitor for fish farmers is naturally the fish population, and many operations are conducted to improve the growth and/or welfare of the fish. 
Usually, the fish population is monitored on a group level using static cameras or passive sensors such as hydrophones or sonars~\cite{Fore:2024:digital_twins}, and parameters such as feeding activity, swimming activity, and size are assessed.
The population can also be monitored on an individual level, for instance with bio-implants~\cite{Svendsen:2021:implant}.
However, the spatial extent of the farms and the large populations make it challenging to obtain an accurate holistic picture of the complete population status. 
As such, underwater vehicles hold an unlocked potential, as they represent moving platforms able to carry a wide array of sensors.  

Research, however, implies that underwater vehicles affect the behavior of farmed salmon and that the fish adjust their swimming pattern to maintain a distance from ROVs~\cite{Kruusmaa:2020:fishmachine_interaction}.
This may indicate that underwater vehicles stress the fish, which has negative health consequences~\cite{ESPELID:1996:salmon_stress}. 
To understand this relationship better, we have conducted field trials aimed at quantifying the fish's behavioral responses to different influence factors. 
In~\cite{zhang:2024:fish_machine}, we installed objects of various shapes and colors in SINTEF ACE's facilities and recorded sonar and image data of the fish. 
Using deep-learning methods, we calculated the fish's avoidance distance to the objects, from which we could conclude that in the trials, the fish kept greater distances to large objects, and to yellow objects versus white objects, and that the object shape had no apparent effect.
Further, the avoidance distance grew proportional to the fish weight. 
In our latest trials, we explored similar fish behavioral responses to an ROV fitted with sonars, hydrophones, and cameras (Fig.~\ref{fig:rov_fish_interaction}).
This data is still under processing.
The experiments represent early steps to increase the knowledge of the dynamics between fish and vehicles in aquaculture, which should govern basic guidelines for operations in environments where fish and vehicles must coexist.

\section{Lessons learned}\label{sec:lessons}
Over the years, we have had our fair share of successes and failures from which we have gathered experience. 
In this section, we will discuss some of the lessons we have learned.

\begin{figure}[t]
    \centering
    \includegraphics[width=0.8\columnwidth]{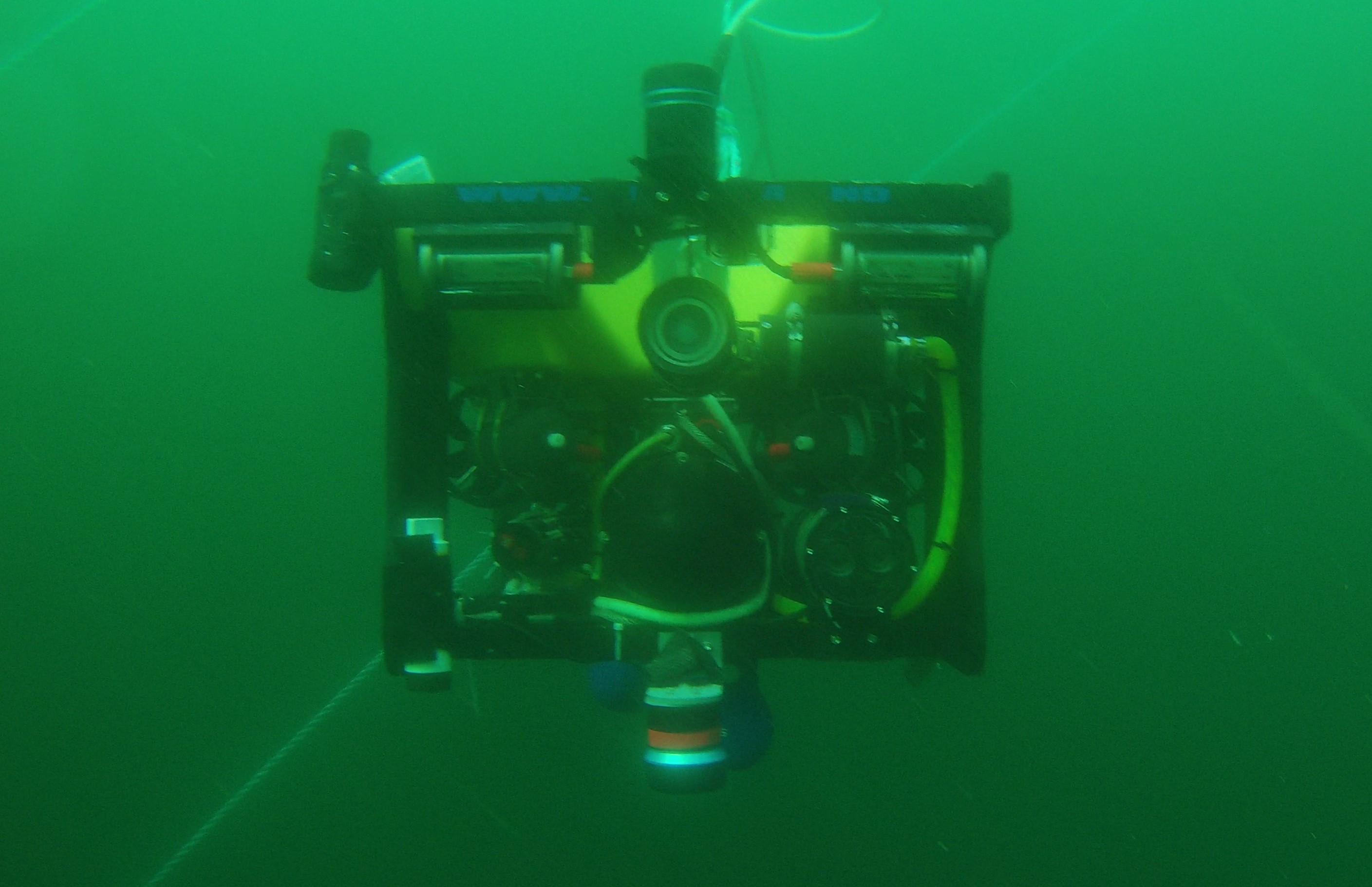}
    \caption{The Argus Mini ROV fitted with sonars, hydrophones, and cameras during trials aimed at identifying behavioural responses of fish to ROV operations. }
    \label{fig:rov_fish_interaction}
\end{figure}

Underwater engineering remains challenging due to the harsh and unforgiving nature of the environment.
Particularly, control is challenging due to hydrodynamics, communication capabilities are limited due to the high signal attenuation in water, and seawater corrosion may break equipment. 
Aquaculture engineering is also exposed to weather conditions (Fig~\ref{fig:snow}), which can increase health, safety and environment (HSE) risks~\cite{Holen2018} and reduce the weather window where operational conditions are considered acceptable.

We've had persistent challenges with acoustic position systems such as USBL.
Specifically, the accuracy and the dropout rate of the measurements have varied between experiments; from experiments where the performance has been well within the error tolerances of the system, to experiments where it has been challenging to get converging measurements.
These problems are likely due to multipath propagation and scattering effects. 
We have learned that USBL system settings should be tuned relative to the environment, and that it is often best to start with a lower transmit power, and then tune until an acceptable signal-to-noise ratio (SNR) is achieved.
Still, frequent and long dropout periods can occur, so state estimators must have acceptable performance during dead-reckoning.

Our experience with DVL has shown that it is able to accurately measure velocity and distance relative to the net.
By setting a high transmit power, net-lock will be achieved when the FOV is unobstructed.
When DVL measurements are intercepted by fish schools during inspections, outliers are generally easy to detect and reject, and dropout periods will usually be short.
Bottom-lock can only be obtained outside of a net cage, and net-lock can only be obtained when reasonably close to the net (typically closer than 5~m). 

Experiments have also shown that water turbidity and lighting conditions have considerable variations in net cages and can highly affect camera images.
In aquaculture, the water turbidity can be quite high due to feed spills and feces from the fish, while operations near the surface may be strongly affected by daylight. 
One specific time this gave us problems was when testing a localization method based on laser-camera triangulation. 
We installed front-facing lasers and then estimated the net-relative pose from the reflection of the lasers seen in the camera images~\cite{Bjerkeng2021}.
In the first trial, we captured data, and post-processing the data yielded very good estimations. 
In the next trials, we aimed to test this method in closed-loop control but were unsuccessful as we were unable to see the laser reflections due to light attenuation in the water.

Tether management is a difficult task, particularly when navigating outside of the net.
Fish farms have complex mooring systems, and power lines and feeding lines also connect the net cages with the feeding barge. 
It is possible to perform inspections outside of the net, but tether management will then require constant focus, and it is usually best to split inspection tasks into segments where there are fewer cables and ropes that can cause entanglement. 
We have experienced one case of tether entanglement as seen in Fig,~\ref{fig:entaglement}, which was caused by a sudden increase in current velocity while the ROV was briefly left unsupervised.

\begin{figure}[t]
    \centering
    \includegraphics[width=0.8\columnwidth]{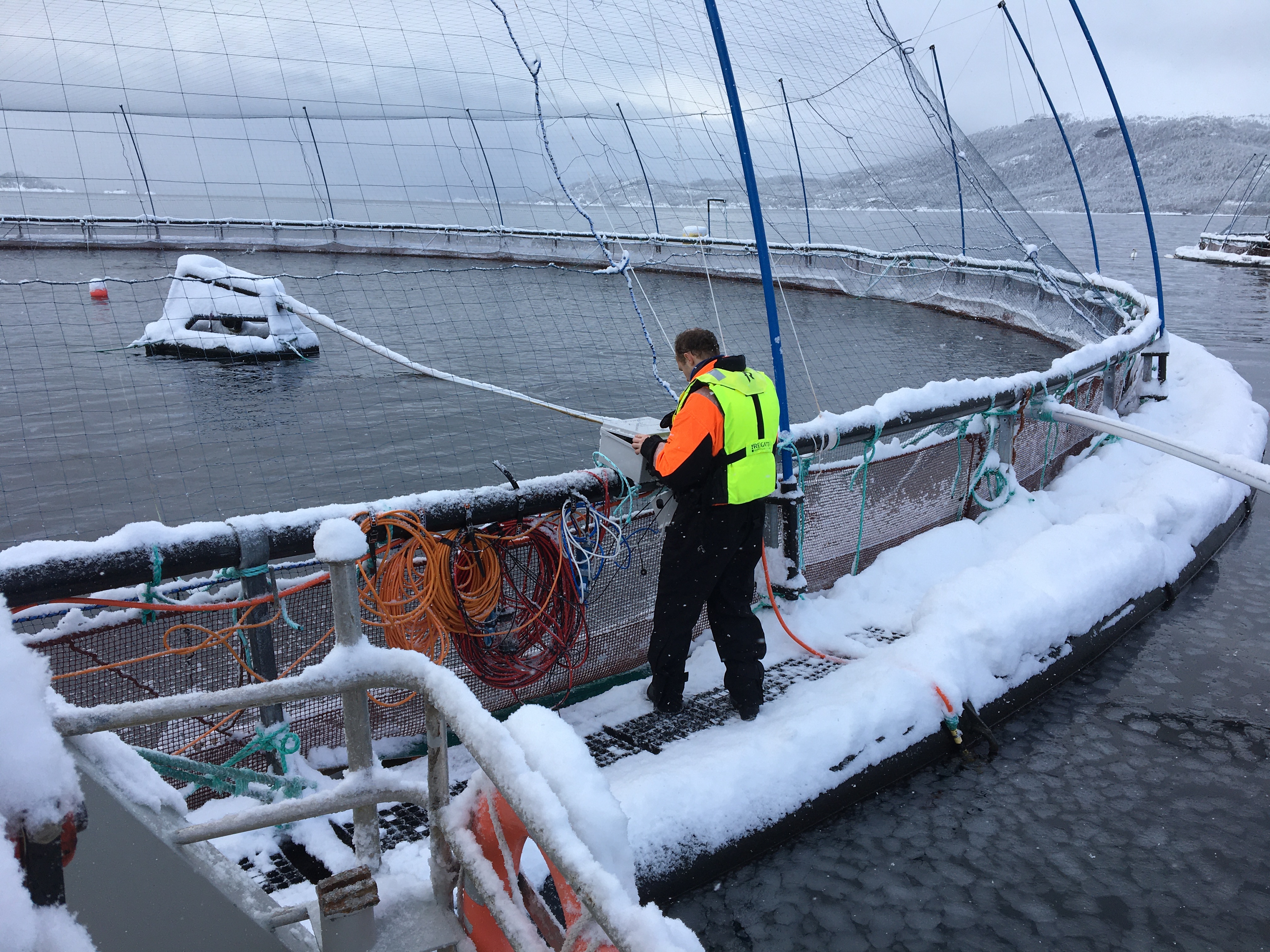}
    \caption{Operations are exposed to weather conditions.}
    \label{fig:snow}
\end{figure}

The risk of entanglement is lower when navigating inside net cages, though navigation on the inside of a net calls for a greater responsibility towards the fish population.
Fast maneuvers can cause flight responses in the fish, which may be related to an increase in stress levels.
Further, we have also experienced that when operating in a specific part of a net cage over time, it appears that the fish population will try to avoid this area. 

Finally, vehicles should be designed in a way that reduces the risk of damaging the net cage structure or harming the fish.
In particular, sharp edges to the vehicle or tools should be avoided as this can cause tear to the net, and thrusters should be covered with lattices to avoid fish getting caught by the propellers. 

\section{Future prospects}\label{sec:future}
In Section~\ref{sec:topics}, we presented various efforts to increase the autonomy level in aquaculture robotics. 
While simple autonomy has been demonstrated, both in the topics discussed in this paper and other papers, more research is needed before human operators can safely be relieved from manually flying ROVs during aquaculture operations. 
Further, the technology readiness level in research has yet to reach a point where it is adopted by the industry. 
Even more, the impact of aquaculture robotics on the fish population has yet to be understood, with the risk that current operations may be harmful to the fish.
Finally, the autonomy level is still far from the point where vehicles can be allowed to safely operate autonomously in fish cages without the possibility for humans to intervene, with the consequence that vehicles still have to remain tethered, thus risking tether entanglement. 

One of the prevailing challenges in aquaculture robotics remains localization and mapping.
While various simultaneous localization and mapping (SLAM) algorithms have been successfully demonstrated in underwater environments, it is challenging to adopt these approaches to aquaculture due to the poor hydroacoustic conditions, the frequent visual occlusions from the fish population, poor visibility due to water turbidity from particles such as feed spills and feces, and the non-static surroundings. 
Methods implemented for aquaculture (Section~\ref{topic:localization}) have demonstrated the ability to estimate the position relative to the net structure, either through sensors attached to the vehicle or the net structure, but no method has yet been able to map the entire complexity of a net cage, which also should include obstacles such as instrumentation, cables, and ropes. 
During net inspections, the FOV of the vehicle is rarely aligned with the direction of the vehicle, which dictates that the vehicle also must maintain an awareness of potential obstacles in its path which is not captured by front-facing cameras or sensors. 

\begin{figure}[t]
    \centering
    \includegraphics[width=0.8\columnwidth]{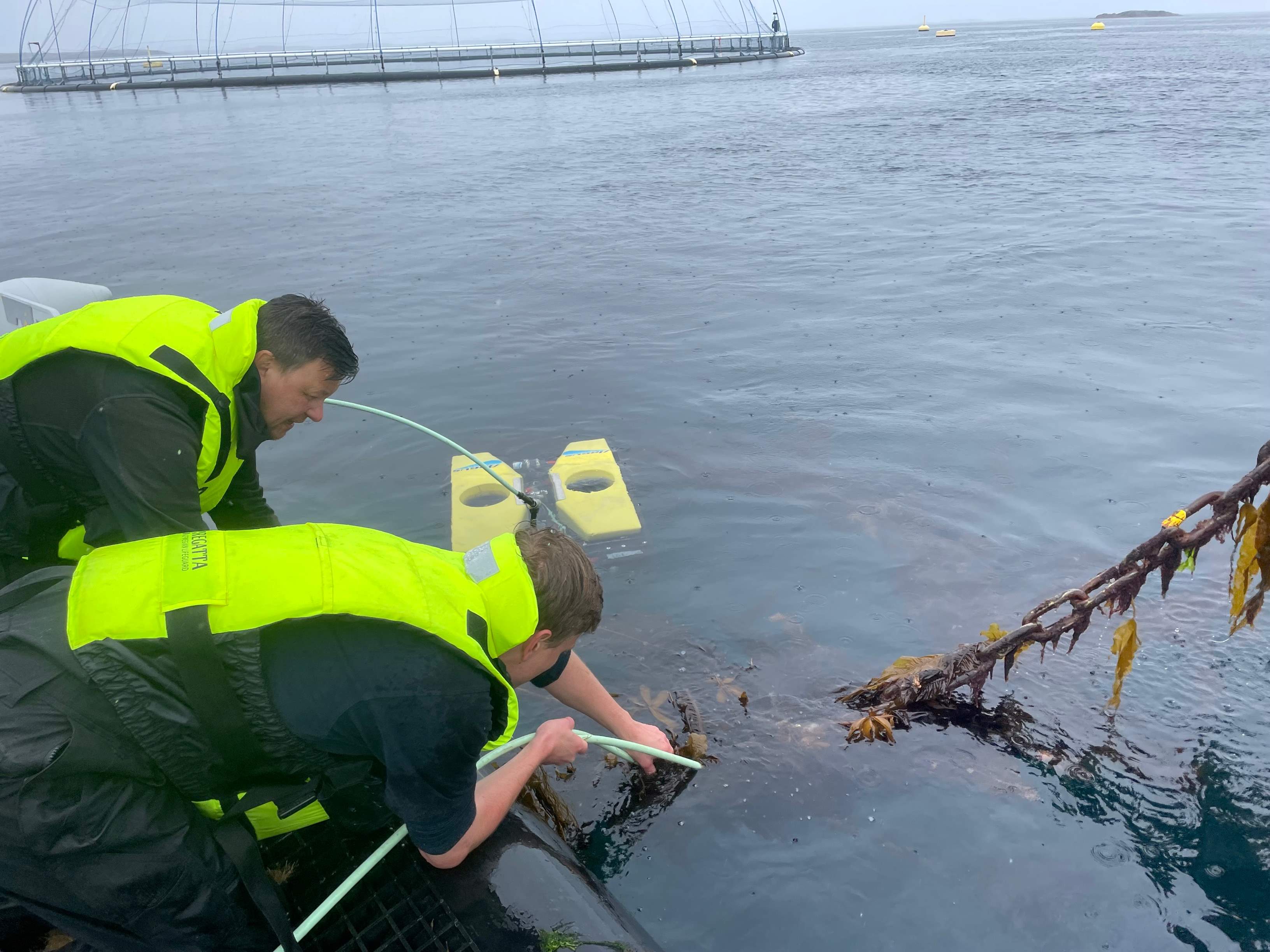}
    \caption{The article author pictured attempting to solve a case of tether entanglement.}
    \label{fig:entaglement}
\end{figure}

SLAM methods for aquaculture also holds great potential for improving reporting after ROV inspections of net cages. 
The current standard is that ROV service companies report inspection results to the fish farmers through a written report and video recording from the ROV camera. 
Research has shown, however, that pilots are not always able to identify structural failures, either due to poor visibility or incomplete coverage~\cite{fiskdir:2020:rov_inspeksjon}. 
Mapping techniques such as photogrammetry may be a valuable tool for improving reporting, making it easier to assess operations and detect structural failures post-mission.

Similar to other industries, underwater vehicle operations remain expensive in aquaculture.
There are many drivers to this; expensive vehicles, expensive sensors, and, most importantly, costs related to the team of operators needed.
A current trend in underwater robotics is the introduction of new commercial vehicles and sensors with significantly lower costs compared to previous standards, which help bring the overall costs down.
Another trend that is bringing costs further down is remote operations where the ROV pilots operate from a land-based facility and control the ROV by wireless transfer of telemetry and video~\cite{Vasilijevic:2023:remote}. 
In the future, we may see more examples of this, which can be combined with an increased autonomy level and permanent resident ROVs at fish farms, removing the need to bring operators or equipment to the farms during missions. 
Realization of permanent resident robots in fish farms may also require development of docking and communication solutions tailored for aquaculture~\cite{kelasidisvendsen2022}. 


Aquaculture robotics have been inspired by underwater robotics in other industries, such as oil and gas, as the technology level of these industries has been more advanced. 
However, aquaculture is an inherently different environment, and, as such, it is not necessarily straightforward to adopt solutions from other industries.
Especially, the interaction between robotic operations and the fish population is poorly understood. 
Further research is required to understand this relationship better, 
such that future aquaculture robotics have better vehicle designs and operational guidelines that are more friendly towards the fish.

In Norwegian aquaculture, governmental incentives are encouraging the development of new production concepts, such as offshore aquaculture, submersible net cages, and rigid and closed structures~\cite{MoeFore:2022:new_concepts}. 
These concepts have fundamental differences compared to the net cages that dominate the industry today.
As such, operations will have to be adapted to these new concepts, which is expected to also affect aquaculture robotics in the future.

Finally, due to the multiple mooring lines, ropes, and cables present at fish farms, tether management remains a difficult task.
If the autonomy level can be increased to a point where the vehicle can safely operate without human surveillance, the tether can be removed, which would represent a milestone in aquaculture robotics that would mitigate completely the risk of entanglement. 
Due to the high safety requirements in aquaculture, this requires a higher level of autonomy than the current standards, as well as rigorous experimental validation.

\section{Conclusion}\label{sec:conclusion}
This paper has presented applications in aquaculture robotics, including localization, planning, control, and robotic interaction with the fish population, with a special focus on field experiments.
These applications showcase potentials and challenges in aquaculture robotics. 
Further, lessons learned from the field are presented, and future prospects and directions in aquaculture robotics are discussed. 

\section*{Acknowledgement}
The results discussed in this paper have been collected through various project and funding schemes: projects funded by the Research Council of Norway (RCN) (pr. numbers: 217541, 256241, 269087, 313737, 327292), the SFI Exposed Center for Research-based Innovation funded by RCN, and internal funding through the SINTEF RACE funding scheme. 
We are grateful to the personell of SINTEF ACE for their help during experiments and to all our collaborators.

\bibliographystyle{IEEEtran}
\bibliography{bibliography}

\end{document}